\documentclass{sig-alternate-05-2015}
\usepackage{makecell}
\usepackage{subfigure}
\usepackage{graphicx}
\usepackage{wrapfig}
\usepackage{colortbl}
\usepackage{balance}
\definecolor{mygray2}{gray}{.8}
\usepackage{amssymb}
\usepackage{pifont}% http://ctan.org/pkg/pifont
\usepackage{colortbl}
\usepackage{hyperref}
\hypersetup{
  colorlinks   = true,    % Colours links instead of ugly boxes
  urlcolor     = blue,    % Colour for external hyperlinks
  linkcolor    = blue,    % Colour of internal links
  citecolor    = red      % Colour of citations
}
\begin{document}

% Copyright 
\setcopyright{acmcopyright}
\setcopyright{acmlicensed}
\setcopyright{rightsretained}
\setcopyright{usgov}
\setcopyright{usgovmixed}
\setcopyright{cagov}
\setcopyright{cagovmixed}
\newcommand*{\addheight}[2][.5ex]{%
  \raisebox{0pt}[\dimexpr\height+(#1)\relax]{#2}%
}
\newcommand{\xmark}{\ding{53}}%
\newcommand{\pic}[1]{\rule{#1}{#1}}% Fake picture of size #1 x #1
\renewcommand\thesubfigure{(\alph{subfigure})}% Allow sub-figure reference to be correctly printed

\newcommand{\etal}{\textit{et al}. } 
\newcommand{\ie}{\textit{i}.\textit{e}., }
\newcommand{\eg}{\textit{e}.\textit{g}., }
\renewcommand{\arraystretch}{1.4}
% DOI

\CopyrightYear{2016} 
\setcopyright{acmcopyright}
\conferenceinfo{MM '16,}{October 15-19, 2016, Amsterdam, Netherlands}
\isbn{978-1-4503-3603-1/16/10}\acmPrice{\$15.00}
\doi{http://dx.doi.org/10.1145/2964284.2967219}
\clubpenalty=10000 
\widowpenalty = 10000
% --- End of Author Metadata ---
\title{Families in the Wild (FIW): Large-Scale Kinship Image Database and Benchmarks}
%\title{Families in the Wild (FIW): }
%\subtitle{A Large-scale Kinship Recognition Database}
\numberofauthors{1} %  in this sample file, there are a *total*
\author{
% 1st. author 
% 1st. author
\alignauthor
Joseph P. Robinson$^\dag$, Ming Shao$^\dag$, Yue Wu$^\dag$, Yun Fu$^{\dag\ddag}$\\
$\dag\ \text{Department of Electrical \& Computer Engineering, Northeastern University, Boston, USA}$\\
$\ddag\ \text{College of Computer \& Information Science, Northeastern University, Boston, USA}$\\
$\text{\{jrobins1, mingshao, yuewu, yunfu\}@ece.neu.edu}$\\
}%
%\alignauthor
%Joseph P. Robinson$^1$\ \ \ \ \ Ming Shao$^1$\ \ \ \ \ Yue Wu$^1$\ \ \ \ \ Yun Fu$^{1,2}$\\\noindent
%\begin{flushleft}
%${1}\ \text{College of Electrical and Computer Engineering, Northeastern University, Boston, MA 02115}$\\
%${2}\ \text{College of Computer and Information Science, Northeastern University, Boston, MA 02115}$
%\end{flushleft}
%\titlenote{Dr.~Trovato insisted his name be first.}\\
%       \affaddr{Northeastern University}\\
%       \affaddr{Boston, Massachusetts}\\
%       \email{robinson.jo@husky.neu.edu}
% 2nd. author
%\alignauthor
% Ming Shao\\
%\titlenote{The secretary disavows any knowledge of this author's actions.}\\
%       \affaddr{Northeastern University}\\
%       \affaddr{Boston, Massachusetts}\\
%       \email{mingshao@ece.neu.edu}
% 3rd. author
%\alignauthor 
%Yue Wu\\
%\titlenote{This author is the
%one who did all the really hard work.}\\
%       \affaddr{Northeastern University}\\
%       \affaddr{Boston, Massachusetts}\\
%       \email{larst@affiliation.org}
%\and  % use '\and' if you need 'another row' of author names
% 4th. author
%\alignauthor Yun Fu\\
%       \affaddr{Northeastern University}\\
%	%\affaddr{Department of Electrical \& Computer Engineering}\\
%       \affaddr{Boston, Massachusetts}\\
%       \email{yunfu@ece.neu.edu}
%}
% Copyright 
%\setcopyright{acmcopyright}
%\setcopyright{acmlicensed}
%\setcopyright{rightsretained}
%\setcopyright{usgov}
%\setcopyright{usgovmixed}
%\setcopyright{cagov}
%\setcopyright{cagovmixed}
%\newcommand*{\addheight}[2][.5ex]{%
%  \raisebox{0pt}[\dimexpr\height+(#1)\relax]{#2}}
\maketitle
\begin{abstract}
We present the largest kinship recognition dataset to date, Families in the Wild (FIW). Motivated by the lack of a single, unified dataset for kinship recognition, we aim to provide a dataset that captivates the interest of the research community. With only a small team, we were able to collect, organize, and label over 10,000 family photos of 1,000 families with our annotation tool designed to mark complex hierarchical relationships and local label information in a quick and efficient manner. We include several benchmarks for two image-based tasks, kinship verification and family recognition. For this, we incorporate several visual features and metric learning methods as baselines. Also, we demonstrate that a pre-trained Convolutional Neural Network (CNN) as an off-the-shelf feature extractor outperforms the other feature types. Then, results were further boosted by fine-tuning two deep CNNs on FIW data: (1) for kinship verification, a triplet loss function was learned on top of the network of pre-train weights; (2) for family recognition, a family-specific softmax classifier was added to the network.

\end{abstract}

%
%  Use this command to print the description
%
\printccsdesc

% We no longer use \terms command

%\%keywords{Large-Scale Database \& Benchmarks; Visual Kinship Dataset; Kinship Verification; Family Recognition; Deep Learning}
%22, 23, 24, 25],
\section{Introduction}
Automatic kinship recognition in visual media is essential for many real-world applications:~\eg kinship verification \cite{dibeklioglu2013like,fang2010towards,hu2014large,xia2012toward,yan2014discriminative,yan2015prototype,zhou2011kinship,zhou2012gabor}, automatic photo library management \cite{xia2012understanding, Shao2014}, historic lineage and genealogical studies \cite{almuashi2015automated}, social-media analysis \cite{guo2014graph}, along with many security applications involving missing persons, human trafficking, crime scene investigations, and even our overall human sensing capabilities--~ultimately, enhancing surveillance systems used in both real-time (\eg vBOLO \cite{xiong2014person} mission\footnote{vBOLO: joint effort of two DHS Centers of Excellence, ALERT \& VACCINE (\href{http://www.northeastern.edu/alert}{http://www.northeastern.edu/alert})}\footnote{\href{https://web-oup.s3-fips-us-gov-west-1.amazonaws.com/default/assets/File/OUP_COETools_Factsheet_vBOLO_051016_PRINT.pdf}{https://web-oup.s3-fips-us-gov-west-1.amazonaws.com/}}) or offline (\eg searching for a subject in a large \textit{gallery} \cite{crosswhite2016template, wang2015face}). 
Thus, a gallery of imagery annotated with rich family information should yield more powerful multimedia retrieval tools and complement many existing facial recognition systems (\eg FBI's NGI\footnote{\href{https://www.eff.org/deeplinks/2014/04/fbi-plans-have-52-million-photos-its-ngi-face-recognition-database-next-year}{https://www.eff.org/2014/fbi-to-have-52M-face-photos}}). However, even after several years (\ie since 2010~\cite{fang2010towards}) there are only a few vision systems capable of handling such tasks. Hence, kin-based technology in the visual domain has yet to truly transition from research-to-reality.

%we are fundamentally developing video analysis tools, with images being the simpler of the two imagery (i.e., multimedia) to generate ground truth (i..e, a database) for. Nonetheless, this is a step towards our overarching goal of devising technology for video analytics, with Kinship being a part of the human sensing capabilities.

We believe the reason that kinship recognition technology has not yet advanced to real-world applications is two-fold:
\begin{enumerate}
\item Current image datasets available for kinship tasks are not large enough to reflect the true data distributions of a family and their members.
\item Visual evidence for kin relationships are less discriminant than class types of other, more conventional machine vision problems (\eg facial recognition or object classification), as many hidden factors affect the similarities of facial appearances amongst family members.
\end{enumerate}
%, and that being in terms of the size of image/identity/family and annotation/tag information. 
 \begin{figure}[!t]%
\centering
\includegraphics[trim={0 .35cm 0 0cm},clip,width=.96\linewidth,height=2.75cm]{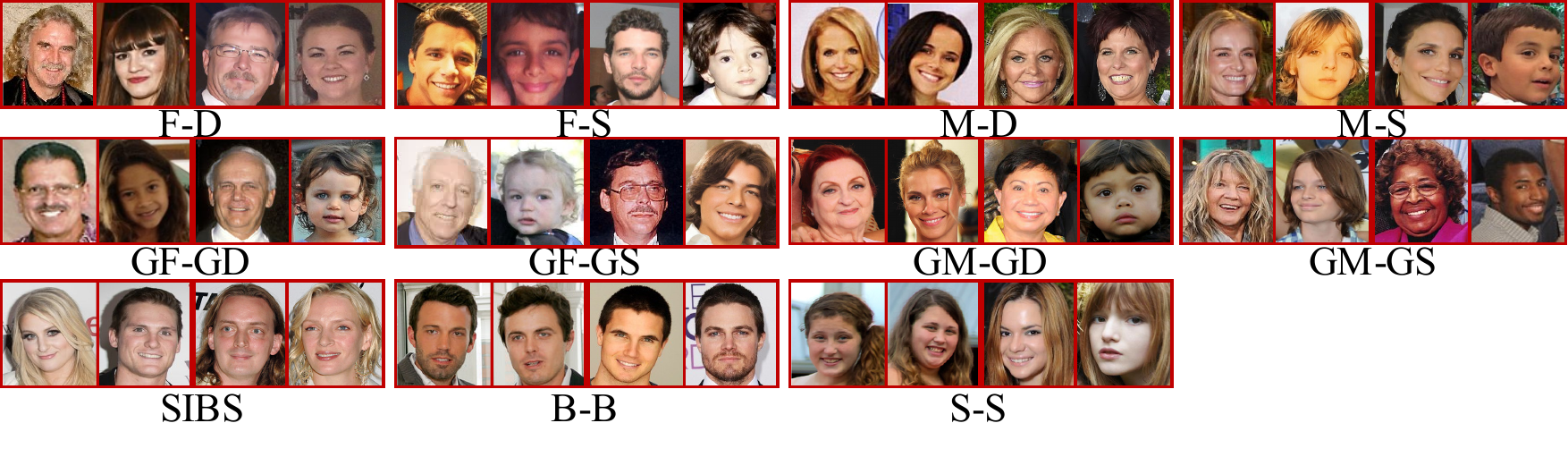}\vspace{-2mm}
\caption{Sample faces chosen of 11 relationship types of FIW. \textit{Parent-child}: (top row) Father-Daughter (F-D), Father-Son (F-S), Mother-Daughter (M-S) Mother-Son (M-S). \textit{Grandparent-grandchild}: (middle row) same labeling convention as above.  \textit{Siblings}: (bottom row)  Sister-Brother (SIBS), Brother-Brother (B-B), Sister-Sister (S-S).}
 \label{fig:pairs} 
\vspace{-3mm}
\end{figure}
\begin{figure*}[t!]
\centering
\includegraphics[width=.93\linewidth, height=.2\textheight]{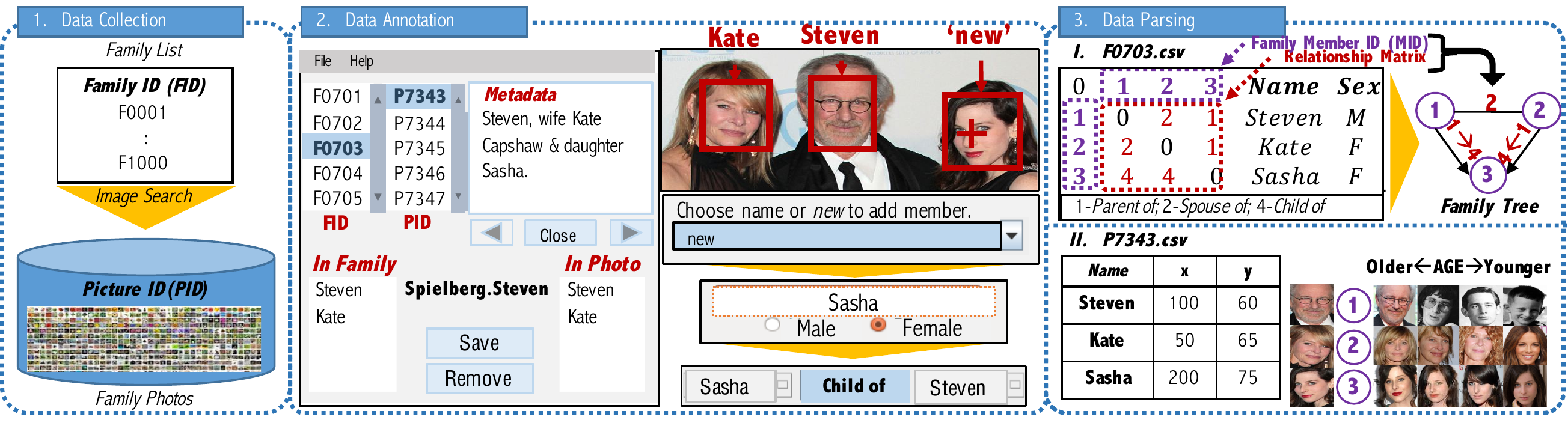}\vspace{-3mm}
   \caption{Method to construct FIW. \textit{Data Collection:} a list of candidate families (with an unique FID) and photos (with an unique PID) are collected. \textit{Data Annotation:} a labeling tool optimized the process of marking the complex hierarchical nature of the 1,000 family trees of FIW. \textit{Data Parsing:} post-processed the two sets of labels generated by the tool to partition data for kinship verification and family recognition.}
\label{fig:annTool}
\vspace{-3mm}
\end{figure*}

To that end, we introduce a large-scale image dataset for kinship recognition called Families in the Wild (FIW). To the best of our knowledge, FIW is by far the \textbf{largest} and \textbf{most comprehensive} kinship dataset available in the vision and multimedia communities (see Table \ref{tbl:compared}). FIW includes $11,163$ unconstrained family photos of $1,000$ families, which is nearly $10x$ more than the next-to-largest, Family101 \cite{fang2013kinship}. Also, it is from these $1,000$ families that $418,060$ image pairs for the $11$ relationship types (see Figure~\ref{fig:pairs} \& Table~\ref{tbl:pair_count1}).

%As listed in Table \ref{tbl:pair_count1}, the 1,000 families provide 418,060 image pairs for the 11 relationship types shown in Figure \ref{fig:pairs}.

Thus far, attempts at image-based kinship recognition have focused on facial features, as is the case in this work. Kinship recognition is typically conducted in one or more of the following modes: (1) kinship verification, a binary classification problem, \ie determine whether or not two or more persons are blood-relatives (\ie kin or non kin); (2) family recognition, a multi-class classification problem, where the aim is to determine the family an individual belongs to; (3) kinship identification, a fine-grain categorization problem, with the goal of determining the types of relationships shared between two or more people. We focus on (1) and (2) in this paper, which we briefly discuss next.

\vspace{1mm}
\noindent\textbf{Kinship Verification.} 
Previous efforts mainly focused on the $4$ parent-child relationship types. As research in psychology and computer vision found, different relationship types render different familiar features, and the $4$ kin relations are usually modeled independently. Thus, it is best to have more kin relationship types accessible-- FIW provides $11$ pair-wise types (see Figure \ref{fig:pairs}): $7$ existing types (\ie parent- child and siblings), but with sample sizes scaling up to $105x$ larger, and $4$ types being offered for the first time (\ie grandparent-grandchild). Also, existing datasets contain, at most, only a couple of hundred pairs per category (\ie $1,000$ in total). Such a insufficient amount leads to models overfitting the training data. Hence, existing models do not generalize well to new, unseen test data. However, FIW now makes $418,060$ pairs available.

%Previous efforts mainly focused on \textit{parent-child} pairs. As research in both psychology and computer vision revealed, different kin relations render different familiar features, and the four kin relations are usually modeled independently. Hence, it is important to have more kin relationship types accessible. Thus, FIW provides 11 pair-wise types [see Figure \ref{fig:pairs}]: 7 returning types  (\ie \textit{parent-child} and \textit{siblings}), but in sample sizes scaling up 105x larger; and 4 types being offered to the research community for the first time (\ie \textit{grandparent-grandchild}).

%In addition, existing datasets contain, at most, only a couple of hundred pairs per category (\ie 1,000 in total). Such a insufficient amount leads to models overfitting the training data. Hence, existing models do not generalize well to new, unseen test data. However, FIW now makes over 418,060 pairs available (as discussed in Section \ref{sec:datastats}).

\vspace{1mm}
\noindent\textbf{Family Recognition.} A challenging task that grows more difficult with more families. This is because families contain large intra-class variations, often overlapping between classes. Similar to conventional face recognition, when the targets are unconstrained \textit{faces in the wild}~\cite{LFWTech} (\ie variation in pose, illumination, expression, and scene) the difficulty level further increases, and the same being true for kinship recognition. These are, unfortunately, challenges that need to be overcome. Thus, FIW poses realistic challenges needed to be addressed before deploying to real-world applications.

\vspace{1mm}
\noindent\textbf{Contributions.} We make three distinct contributions:
\begin{enumerate}
\item We introduce the largest visual kinship dataset to date, Families in the Wild (FIW).\footnote{FIW will be available upon publication of this paper.} 
FIW is complete with rich label information for $1,000$ family trees, from which $11$ kin relationship types were extracted in numbers that are orders of magnitude times larger than existing datasets. This was made possible with an efficient annotation tool and procedure (Section \ref{sec:fw}).
\item We provide several benchmarks on FIW for both kinship verification and family recognition, including various low-level features, metric learning methods, and pre-trained Convolutional Neural Network (CNN) models. See Section \ref{sec:experimental} for experimental settings and results.
\item We fine-tune two CNNs: one with a triplet-loss layer on top, and the other with a softmax loss. Both yield a significant boost in performance over all other benchmarks for both tasks (Section \ref{subsec:verification}).
\end{enumerate}

%
%\begin{table*}[t]
%\begin{center}
%\centering
%\caption{Comparison of FIW with related datasets.}
%	\footnotesize
%	\centering
%	\begin{tabular}{l|p{.4in}p{.4in}p{.35in}p{.34in}p{.55in}p{1.70in}}\hline
%	 Dataset & \footnotesize No. Family& No. People&  No. Faces&  Age Varies & Family Structure & Highlights\\ \hline
%	CornellKin &\ 150 & 300 & 300 & No & No & Parent-child pairs.\\
%	UB KinFace-I&\ 90 & 180 & 270 & Yes & No & Parent-child pairs. Parents' images at various ages.\\
%	UB KinFace-II\cite{lu2015fg}&\ 200 & 400 & 600 & Yes & No & Parent-child pairs. Parents' images at various ages.\\
%	KFW-I&\cite{lu2014kinship}\ --- & 1066 & 1066 & No & No & Parent-child pairs.\\
%	KFW-II&\ --- & 2000 & 2000 & No & No & Parent-child pairs.\\
%	TSKinFace &\ 787 & 2589 &\ --- & Yes & Yes & Two parents-child pairs for tri-verification.\\
%	Family101 \cite{fang2013kinship}&\ 101 & 607 & 14,816 & Yes & Yes & Family structured, variations in age and ethnicity. \\
%	FIW(Ours) &\ 936 & 10,676 & 30,725 & Yes & Yes & A corpus of 1k family trees that provides both depth and breadth and multi-task evaluation offerings.\\ 
%\end{tabular}
% \label{tbl:compared} 
%\end{center}
%\end{table*}
%

\section{Families in the Wild}\label{sec:fw}
We now discuss the procedure followed to collect, organize, and label $11,193$ family photos of $1,000$ families with minimal manual labor. Then, we statistically compare FIW with other related datasets.

\subsection{Building FIW}
The goal for FIW was to collect around $10$ photos for $1,000$ families, each with at least $3$ family members. We now summarize the method for achieving this in a three step process, which is visually depicted in Figure \ref{fig:annTool}.

\vspace{1mm}
\noindent\textbf{Step 1: Data Collection.} A list of over $1,000$ candidate families was made. To ensure diversity, we targeted groups of public figures worldwide by searching for <\textit{ethnicity} OR \textit{country}> AND <\textit{occupation}> online (\eg \textit{MLB} [Baseball] \textit{Players}, \textit{Brazilian Politicians}, \textit{Chinese Actors}, \textit{Denmark + Royal Family}). Family photos were then collected using various search engines (\eg Google, Bing, Yahoo) and social media outlets (\eg Pinterest) to widen the search space. Those with at least 3 family members and $8$ family photos were added to FIW under an assigned Family ID (FID).

\vspace{1mm}
\noindent\textbf{Step 2: Data Annotation.} A tool was developed to quickly annotate a large corpus of family photos.
%\footnote{FIW, along with source code and experimental data (\ie parameters, fine-tuned CNN model, and features) will be made available for download upon publication of this paper.} 
All photos for a given FID are labeled sequentially. Labeling is done by clicking a family member's face. Next, a face detector initializes a resizable box around the face. Faces unseen by the detector are discarded, as these are assumed to be poorly resolved. The tool then prompts for the name of the member via a drop-down menu. For starters, option new adds a member to a family under a unique member ID (MID), which prompts for the name, gender, and relationship types shared with others previously added to the current family (or FID). From there onward, labeling family members is just a matter of clicking.

\vspace{1mm}
\noindent\textbf{Step 3: Dataset Parsing.} Two sets of labels are generated in Step 2: (1) \textit{image-level}, containing names and corresponding facial locations; (2) \textit{family-level}, containing a relationship matrix that represents the entire family tree. Relationship matrices are then referenced to generate lists of member pairs for the 11 relationship types. Next, face detections are normalized and cropped with \cite{dlib09}, then are stored according the previously assigned \textit{FID}$\rightarrow$ \textit{MID}. Lastly, lists of image pairs are generated for both kinship verification and family recognition.
%, each with family trees made-up of relationships that span with both depth and breadth, \ie across multiple generations (\eg grandfather and granddaughter) and within first and second generations \eg, brothers, sisters, aunts, uncles, and even spanning across families of in-laws. 
\subsection{Database Statistics}\label{sec:datastats}

Our FIW dataset far outdoes its predecessors in terms of quantity, quality, and purpose. FIW contains $11,193$ family photos of $1,000$ different families. There are about $10$ images per family that include at least $3$ and as many as $24$ family members. We compare FIW to related datasets in Table \ref{tbl:compared} and  \ref{tbl:pair_count1}. Clearly, FIW provides more families, identities, facial images, relationship types, and labeled pairs-- the pair count of FIW is orders of magnitude bigger than the next- to-largest (\ie KFW-II).

%Our FIW dataset far outdoes its predecessors in terms of quantity, quality, and purpose. FIW consists of over 10,000 family photos of 1,000 families, each with about 10 images containing at least 3 and as many as 24 family members. We compare FIW to related datasets in Table \ref{tbl:compared} and  \ref{tbl:pair_count1}. FIW provides more families, identities, facial images, relationship types, and labeled pairs-- the pair count of FIW is 105 magnitudes larger than the next to largest (\ie KFW-II).

\section{Experiments On FIW}\label{sec:experimental}
In this section, we first discuss visual features and related methods used to benchmark the FIW dataset. We then report and review all benchmark results. Finally, we discuss the two methods used to fine-tune the pre-trained CNN model: (1) training a triplet-loss for kinship verification and (2) learning a softmax classifier for family recognition. Top scores were obtained for both in the respective task.

\begin{table}[t]
\centering
\caption{Comparison of FIW with related datasets.}
  \scriptsize
  \centering
  \begin{tabular}{|l|*{2}{p{.85cm}<{\centering}}*{2}{p{.6cm}<{\centering}}{p{1.2cm}<{\centering}|}}
  % \begin{tabular}{l|p{.33in}p{.37in}p{.3in}p{.3in}p{.53in}}\hline
  \Xhline{.8pt}
  Dataset & No. Family & No. People &  No. Faces &  Age Varies & Family Trees \\ \hline
  CornellKin\cite{fang2010towards} & 150  & 300 & 300 & \xmark & \xmark \\
  UBKinFace\cite{Ming_CVPR11_Genealogical,Xia201144}& 200 & 400 & 600 & \ 
\Large\checkmark
 & \xmark \\
  KFW-I\cite{lu2014neighborhood}& \xmark & 1,066 & 1,066 & \xmark & \xmark\\
  KFW-II\cite{lu2014neighborhood}& \xmark & 2,000 & 2,000 & \xmark & \xmark \\
  TSKinFace\cite{qin2015tri} & 787 & 2,589 & \xmark & \Large\checkmark & \Large\checkmark \\
  Family101\cite{fang2013kinship} & 101 & 607 & 14,816 & \Large\checkmark & \Large\checkmark \\\hline
  \multicolumn{1}{|>{\columncolor{mygray2}}l|}{FIW(Ours)} & \multicolumn{1}{>{\columncolor{mygray2}}c}{\textbf{1,000}} & \multicolumn{1}{>{\columncolor{mygray2}}c}{\textbf{10,676}} & \multicolumn{1}{>{\columncolor{mygray2}}c}{\textbf{30,725}} & \multicolumn{1}{>{\columncolor{mygray2}}c}{\Large\checkmark} & \multicolumn{1}{>{\columncolor{mygray2}}c|}{\large\checkmark} \\
  \Xhline{.8pt}
\end{tabular} 
 \label{tbl:compared}
\vspace{-3mm}
\end{table}
	
	\subsection{Features and Related Methods}
\label{subsec:features}
All features and methods covered here were used to bench- mark FIW. First, we review handcrafted features, Scale In- variant Feature Transformation (SIFT) and Local Binary Patterns (LBP), which are both widely used in kinship verification \cite{lu2014neighborhood} and facial recognition \cite{schroff2015facenet}. Next, we introduce VGG-Face, the pre-trained CNN model used here as an off- the-shelf feature extractor. Lastly, we review other related metric learning methods.

\vspace{1mm}
\noindent\textbf{SIFT~\cite{lowe2004distinctive}} features have been widely applied in object and face recognition. As done in \cite{lu2014neighborhood}, we resized all facial images to $64\times 64$, and set the block size to $16\times 16$ with a stride of $8$. Thus, there were a total of $49$ blocks for each image, yielding a feature vector of length $128\times 49 = 6,272\text{D}$.

\begin{table}[!t]
\begin{center}

\caption{Pair counts for FIW and related datasets.}
\vspace{-2.3mm}
	\scriptsize
	  %\begin{tabular}{|{p{12mm}<{\centering}}p{10mm}p{10mm}p{10mm}p{10mm}{p{10mm}<{\centering}|}}
	  
	\begin{tabular}{|p{12mm}|p{10.4mm}p{10mm}p{10mm}p{10mm}|p{10mm}|}
	%\hline
	\Xhline{.8pt}
%	c>{\columncolor[gray]{0.8}}}
	 &KFW-II \cite{lu2014neighborhood}& Sibling Face~\cite{guo2014graph}& Group Face~\cite{guo2014graph}& Family 101\cite{fang2013kinship}& \cellcolor{mygray2}FIW (Ours)\\ \hline
	B-B &~~~\xmark& ~~232 & ~40 &~~\xmark& \cellcolor{mygray2}\textbf{86,000}\\
	S-S &~~~\xmark& ~~211 & ~32 &~~\xmark& \cellcolor{mygray2}\textbf{86,000}\\
	SIB &~~~\xmark& ~~277 & ~53 &~~\xmark& \cellcolor{mygray2}\textbf{75,000}\\ \hline
	F-D &~~250 &~~~\xmark& ~69 &~147 & \cellcolor{mygray2}\textbf{45,000}\\
	F-S &~~250 &~~~\xmark& ~69 &~213 & \cellcolor{mygray2}\textbf{43,000}\\
	M-D &~~250 &~~~\xmark& ~62 &~148 & \cellcolor{mygray2}\textbf{44,000}\\
	M-S &~~250 &~~~\xmark& ~70 &~184 & \cellcolor{mygray2}\textbf{37,000}\\ \hline
	GF-GD&~~~\xmark&~~~\xmark&~\xmark&~~\xmark& \cellcolor{mygray2}~~\textbf{410}\\
	GF-GS&~~~\xmark&~~~\xmark&~\xmark&~~\xmark& \cellcolor{mygray2}~~\textbf{350}\\
	GM-GD&~~~\xmark&~~~\xmark&~\xmark&~~\xmark& \cellcolor{mygray2}~~\textbf{550}\\
	GM-GS&~~~\xmark&~~~\xmark&~\xmark&~~\xmark& \cellcolor{mygray2}~~\textbf{750}\\ 
	\hline
	Total & ~1,000 & ~~720 &395 &~607 &\cellcolor{mygray2}\textbf{418,060}\\ \Xhline{.8pt}
\end{tabular}
\label{tbl:pair_count1}
\end{center}
\vspace{-3mm}
\end{table}

 \begin{figure*}[!t]%
\centering
\includegraphics[trim={3.1mm .35cm 3.1mm 0cm},clip,width=.8\linewidth,height=7.5cm]{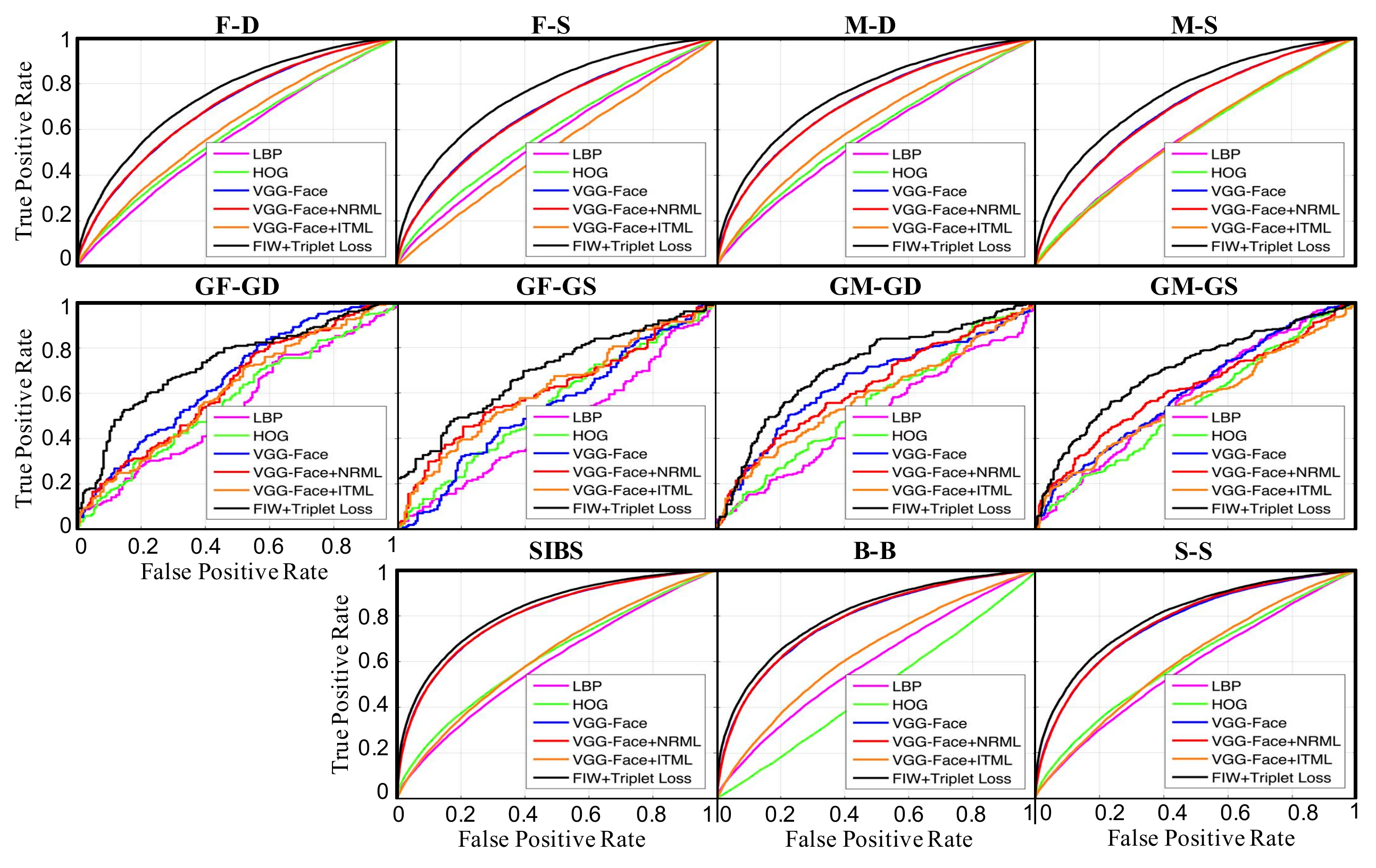}\vspace{-3mm}
\caption{Relationship specific ROC curves depicting performance of each method.}
 \label{fig:roc}
\vspace{0mm}
\end{figure*}

\begin{table*}[h!t]

\begin{center}
\caption{Verification accuracy scores (\%) for 5-fold experiment on FIW. No family overlap between folds.}
	\scriptsize
	\begin{tabular}{|l|cccccccccccc|} \Xhline{0.8pt}
	\label{tab:KinshipVer}
	 		&F-D		&	F-S	&	M-D	&	M-S	&	SIBS	&B-B&S-S&	GF-GD	&GF-GS&	GM-GD&	GM-GS&Avg.\\\hline
	 SIFT	&56.1&56.5&56.4&55.3&58.7&50.3&57.4&59.3&66.9&60.4&56.9&57.7$\pm$4.0\\\hline 
	 
	 LBP	&55.0&55.3&55.4&55.9&57.1&56.8&55.8&58.5&59.1&55.6&60.1&56.8$\pm$1.7 \\\hline
	VGG-Face&64.4&63.4&66.2&64.0&73.2&71.5&70.8&64.4&68.6&66.2&63.5&66.9$\pm$3.5 \\\hline 	  		
	Fine-Tuned CNN&\textbf{69.4}& \textbf{68.2}& \textbf{68.4}& \textbf{69.4}& \textbf{74.4}& \textbf{73.0}& \textbf{72.5}& \textbf{72.9}& \textbf{72.3}& \textbf{72.4}& \textbf{68.3} &\textbf{71.0$\pm$2.3} \\\hline	 
	% Avg.	&58.6	&58.7	&59.6&58.4	&\textbf{63.7}	&62.7&61.9&--&--&--&--&--\\
\end{tabular}
\end{center} 

\vspace{0mm}
\end{table*}

\vspace{1mm}
\noindent\textbf{LBP~\cite{ahonen2006face}} has been frequently used for texture analysis and face recognition, as it describes the appearance of an image in a small, local neighborhood around a pixel. Once again, we followed the feature settings of \cite{lu2014neighborhood} by first resizing each facial image to $64\times 64$, and then extracting LBP features from $16\times 16$ non-overlapping blocks with a radius of 2 pixels and number of neighbors (\ie samples) set to 8. We then generated a 256D histogram from each, yielding a final feature vector of length $256\times 16 = 4,096\text{D}$.

\vspace{1mm}
\noindent\textbf{VGG-Face CNN \cite{Parkhi15}} uses a ``Very Deep'' architecture with very small convolutional kernels (\ie $3\times 3$) and convolutional stride (\ie $1$ pixel). This model was pre-trained on over $2.6$ million images of $2,622$ celebrities. For this, each face image was resized to $224\times 224$ and then fed-forward to the second-to-last fully-connected layer (\ie fc7) of the CNN model, producing a $4,096$D feature vector.

\vspace{1mm}
\noindent\textbf{Metric Learning} methods are commonly employed in the visual domain, some designed specifically for kinship recognition \cite{dehghan2014look, lu2014neighborhood,qin2015tri} and some as generic metric learning methods \cite{davis2007information}. We chose two representative methods from these two categories to report benchmarks on, Neighborhood Repulsed Metric Learning (NRML) \cite{lu2014neighborhood} and Information Theoretic Metric Learning (ITML) \cite{davis2007information}.

\subsection{Fine-Tuned CNN Model}\label{subsec:verification}
Weights of deep networks are trained on larger amounts of generic source data, then fine-tuned on target data, which utilize a wider, more readily available source domain that resembles the target in either modality, view, or both \cite{DBLP:journals/corr/GirshickDDM13}. Following this notion, and motivated by recent success with deep learning on faces \cite{Parkhi15, sun2014deep, wang2015deep}, we fine-tuned the VGG-Face model, improving results for both kinship verification and family recognition.

\vspace{1mm}
\noindent\textbf{Kinship Verification.} 
In kinship verification, for each fold we select the families which have more than $10$ images in the rest four folds. $90\%$ images of each family are used to fine-tune the model and the rest for validation. An average of $8,295$ images are selected. The experimental results of this part can be found in Table~\ref{tab:KinshipVer} and Figure~\ref{fig:roc}. This is so far the best results on FIW. Specifically, we remove the last fully-connected layer which is used to identify $2,622$ people and employed a triplet-loss \cite{schroff2015facenet} as the loss function. The second-to-last fully-connected layer was the only non-frozen layer of the original CNN model, with an initial learning rate of $10^{-5}$ that decreased by a factor of $10$ every $700$ iterations (out of $1,400$). Batch size was $128$ images, and other network settings were the same as the original VGG-Face model. Training was done on a single GTX Titan X with about 10GB GPU memory. Fine-tuning was done using the renown Caffe \cite{jia2014caffe} framework.

\vspace{1mm}
\noindent\textbf{Family Recognition.} 
As visual kinship recognition is essentially related to facial features, we froze the weights in the lower levels of the VGG-Face network, and replaced the topmost layer with a new softmax layer to classify the $316$ families. Other settings are the same as those used for kinship verification.

%    Family Recognition 5-fold experimental results. To fine-tune on FIW we replaced the top-layer of the VGG-Face with a softmax layer learned for 316 families (\ie 316D output).}%\label{tab:FamilyRecFineTune}  
\begin{table}[!b]

\vspace{-3mm}
\centering
    \caption{Family Recognition accuracy scores (\%) for 5-fold experiment on FIW (316 Families). No family overlap between folds.}
\scriptsize
\label{tab:famclass}
\begin{tabular}{lcccccc}
\Xhline{.8pt}
& 1 & 2 & 3 & 4 &  5 & Avg.\\
\hline
VGG & 9.6 & 14.5 & 11.6 & 12.7 & 13.1 & 12.3$\pm$1.8 \\ 
Fine-tuned & \textbf{10.9} & \textbf{14.8} & \textbf{12.5} & \textbf{14.8} & \textbf{13.5} & \textbf{13.3$\pm$1.6}\\
  \Xhline{.8pt}
  
\end{tabular}
\end{table}

\subsection{Experimental Settings}
In this section, we provide benchmarks on FIW using all features and related methods mentioned in the previous section. Dimensionality of each feature is reduced to 100D using PCA. Experiments were done following a 5-fold cross-validation protocol. Each fold was of equal size and with no family overlap between folds.

\vspace{1mm}
\noindent\textbf{Experiment 1: Kinship Verification.} We randomly select an equal number of positive and negative pairs for each fold family. Cosine similarity is computed for each pair in the test fold. The average verification rate of all folds is reported in Table \ref{tab:KinshipVer}, showing that kinship verification is a challenging task. While some relation are relatively easy to recognize, \eg B-B, SIBS, S-S through SIFT, LBP, and VGG-Face features, results of other relations such as parent- child are still below 70.0\%. Clearly, VGG-Face features are much better than hand-craft features. Notice grandparent- child pairs typically have higher accuracies than parent-child, which we believe is due to the differences in sample sizes. We also compare with the state-of-the-art metric learning methods, NRML and ITML (see Figure \ref{fig:roc}). Showing improved scores for the low-level features, but still outperformed by the CNN model fine-tuned on FIW.

\vspace{1mm}
\noindent\textbf{Experiment 2: Family Recognition.} We again follow the 5-fold cross-validation protocol with no family overlap. Families with 6 or more members, from which the 5 members with the most images were used. The results in Table \ref{tab:famclass} are from 316 families with 7,772 images. Folds were made up of one member for each family. Multi-class SVM was used to model VGG-Face features for each family (\ie one-vs-rest). We then improved the top-1 classification accuracy from 12.3$\pm$1.8 (\%, VGG-Face) to 13.3$\pm$1.6 (\%, our fine-tuned model).

%\vspace{-2mm}
\section{Discussion}
%We introduced a large-scale dataset of family photos captured in natural environments and in unconstrained settings (\ie Families in the Wild). An annotation tool was designed to quickly generate rich label information for over 10,000 photos of 1,000 families. Emphasis was put on diversity (\ie families worldwide), data distribution (\ie at least 3 members and 8 photos per family), sample sizes (\ie multiple instances of each member, and at various ages), quality (\ie only faces seen by the detector), and quantity (\ie much more data and new relationship types).
%
%There are many interesting directions for future work on our dataset. We currently only verify whether a pair of images is kin or non kin, however, also predicting the type of kin relationship may lead to more value and practical usefulness. Thus, bringing us closer to doing fine-grain categorization on entire family trees. FIW will be an ongoing effort that will continue to grow and evolve. See project page for downloads, updates, and to learn more about the FIW dataset.
%
We introduced a large-scale dataset of family photos captured in natural, unconstrained environments (\ie Families in the Wild). An annotation tool was designed to quickly generate rich label information for $11,163$ photos of $1,000$ families. Emphasis was put on diversity (\ie families worldwide), data distribution (\ie at least 3 members and 8 photos per family), sample sizes (i.e., multiple instances of each member, and at various ages), quality (\ie only faces seen by the detector), and quantity (\ie much more data and new relationship types).

There are many interesting directions for future work on our dataset. We currently only verify whether a pair of images is kin or non kin. However, also predicting the kin relationship type could lead to more value and practical usefulness. Thus, bringing us closer to doing fine-grain categorization on entire family trees. FIW will be an ongoing effort that will continually grow and evolve. See project page for downloads, updates, and to learn more about FIW.

%\pagebreak
%\pagebreak

\noindent\textbf{Acknowledgements}\\
%\section*{Acknowledgements}
This material is based upon work supported by the U.S. Department of Homeland Security, Science and Technology Directorate, Office of University Programs, under Grant Award 2013-ST-061-ED0001. The views and conclusions contained in this document are those of the authors and should not be interpreted as necessarily representing the official policies, either expressed or implied, of the U.S. Department of Homeland Security. 

We would also like to thank all members of SMILE Lab who helped with the process of collecting and annotating the FIW dataset.

%ACKNOWLEDGMENTS are optional
%
%the
%authors would like to thank Gerald Murray of ACM for
%his help in codifying this \textit{Author's Guide}
%and the \textbf{.cls} and \textbf{.tex} files that it describes.

\bibliographystyle{abbrv}
\bibliography{sigproc}  % sigproc.bib is the name of 
\end{document}